# Lightweight Probabilistic Deep Networks

Jochen Gast    Stefan Roth

Department of Computer Science, TU Darmstadt

## Abstract

*Even though probabilistic treatments of neural networks have a long history, they have not found widespread use in practice. Sampling approaches are often too slow already for simple networks. The size of the inputs and the depth of typical CNN architectures in computer vision only compound this problem. Uncertainty in neural networks has thus been largely ignored in practice, despite the fact that it may provide important information about the reliability of predictions and the inner workings of the network. In this paper, we introduce two lightweight approaches to making supervised learning with probabilistic deep networks practical: First, we suggest probabilistic output layers for classification and regression that require only minimal changes to existing networks. Second, we employ assumed density filtering and show that activation uncertainties can be propagated in a practical fashion through the entire network, again with minor changes. Both probabilistic networks retain the predictive power of the deterministic counterpart, but yield uncertainties that correlate well with the empirical error induced by their predictions. Moreover, the robustness to adversarial examples is significantly increased.*

## 1. Introduction

In recent years, deep convolutional networks have become the workhorse for many applications, such as image classification [20], object detection [15], semantic labeling [37], or optical flow estimation [10]. While details of the employed architectures differ, there is typically the common notion that activations and predictions are represented as *point estimates*. In practice, this means that most architectures do not exhibit an explicit representation of uncertainty – network predictions are agnostic of whether they are reliable or not. Even worse, predictive distributions of common classification models relying on the softmax are not well calibrated and tend to be overconfident [4, 13, 19]. Additionally, they can be easily fooled even with imperceptible changes made to the input image [40]. While the focus on application performance may be desired from the perspec-tive of benchmark comparisons, there are many other real-world settings where the robustness and reliability of predictive distributions is much more important, *e.g.* in medical applications [52] or autonomous driving [2]. Here, uncertainties are crucial, since they enable us to treat highly uncertain predictions with particular care. It is interesting to note that prominent cases where modern AI systems fail increasingly make headlines in mainstream media, *e.g.* [9].

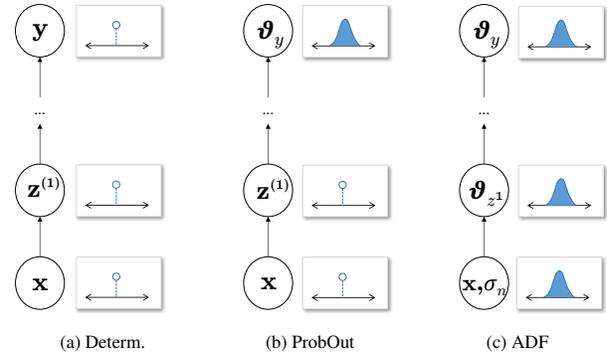

Figure 1. **Uncertainties in CNNs**: (a) Traditional deep networks represent both activations and outputs as deterministic point estimates. (b) In this work, we explore the replacement of outputs by probabilistic output layers. (c) To go one step further, we also consider replacing all intermediate activations by distributions.

More recently, interest in *becoming more Bayesian* in deep learning has picked up momentum, *c.f.* [5, 6]. Bayesian methods, despite being principled, often complicate inference [18], or require expensive test-time sampling [13]. In this work, we extend deep neural networks to probabilistic predictions, while at the same time making minimal changes to existing network architectures and keeping inference fast and efficient. Hence, we propose to not "go down the full Bayesian road", but obtain uncertainties in a *lightweight manner*, which can be applied to well-proven networks. In a first step, we rely on *probabilistic output layers*, termed ProbOut for brevity, which replace standard point predictions from deterministic networks by distributions over the output (Fig. 1, a–b). Subsequently, we go one step further by replacing intermediate activations by distributions as well (Fig. 1, b–c). Building on classical Gaussian belief networks [12], we use assumed density filtering (ADF) [7] to propagate activation uncertainties through the network in a single pass. In contrast to full Bayesian networks, which treat parameters probabilistically, we only replace the activations, enabling us to keep existing networks without a significant increase in the number of parameters.

Concretely, we propose *general power exponential outputs* for regression problems, which we demonstrate with an optical flow application using FlowNet [10]. For classification problems, we suggest a *Dirichlet layer*, which allows assessing the uncertainty of class predictions. Both approaches are trained by standard conditional maximum likelihood estimation. The resulting probabilistic networks keep the predictive power and efficiency of the deterministic counterpart, but know when their predictions are not reliable. Moreover, the probabilistic treatment renders classification networks more resilient to adversarial attacks [17].

## 2. Related Work

**Latent generative representations.** Perhaps among the first probabilistic neural networks are the works relating undirected graphical models to neural networks via Boltzmann machines [3, 24]; similarly, links are established for deep belief nets (DBN) via sigmoid belief nets [23, 41]. Later, [12, 22] introduce Gaussian units to represent latent variables that are nonlinearly propagated through network layers. Note that these classic forms of directed or undirected networks essentially recover hidden, generative representations of the data. A more recent approach to latent, generative modeling is the reparametrization trick used in variational autoencoders [32], or conditional autoencoders [30, 46]. In this work, we do not consider generative modeling, but address supervised learning tasks that are widely used in vision, *e.g.* [10, 15, 37]. However, we build upon the layerwise transformations of Gaussian distributions by Hinton and Frey [12] in the context of nonlinear Gaussian belief networks, and repurpose them here for the representation of *activation uncertainties* in deep supervised learning.

**Bayesian approaches.** One way towards probabilistic predictions is the Bayesian treatment of parameters, *e.g.* recovering tractable distributions of weights [5, 14, 18, 21, 44]. While a Bayesian treatment recovers model uncertainties in a principled way, it also increases the number of parameters, as weights are represented by means of a parametric model. Monte Carlo estimates are commonly computed, *e.g.* to approximate variational bounds, increasing the computational burden at test time. To address this, Korattikara *et al.* [33] train a student to reproduce the behavior of a teacher network trained in the full Bayesian setting. We do not require separate networks and require only minor changes to well-established procedures for loss-based supervised training.

**Sampling-based approaches.** Bouchacourt *et al.* [6] introduce DISCO networks, that, given injected noise, minimize the dissimilarity coefficient between the network model distribution and the true data distribution. Lakshminarayanan *et al.* [35] obtain predictive uncertainties from an ensemble of $M$ networks; however, while training multiple networks may be viable for small architectures, it is not practical for large networks used, *e.g.*, in vision, neither at training nor at test time. Gal *et al.* [13] train networks under Bernoulli units, which are then kept at test-time to compute Monte Carlo estimates of the uncertainties. One limitation is the possibility to construct examples, where test-time dropout does not calibrate its predicted uncertainty to the observed variance in the data [43]. [13] can be combined with variational dropout [31] to improve predictions by effectively learning dropout rates from data.

Kendall *et al.* [28] combine model uncertainty from Monte Carlo sampling, termed epistemic uncertainty, with noise from the observation model, termed aleatoric, heteroscedastic uncertainty. In this terminology our approach falls under the category of *aleatoric, heteroscedastic uncertainty*. Kendall *et al.* argue that for many vision tasks model uncertainty is less important than uncertainty from observation noise, since it is a source of entropy that cannot be explained away by more evidence. Hence, in this work we disregard model uncertainty and aim for a highly practical and fast approach to treating aleatoric uncertainty.

Common to methods that sample at test time to obtain uncertainties is that, at least currently, they are too slow for a number of important applications, *e.g.* autonomous driving.

**Uncertainty propagating architectures.** Abdelaziz *et al.* [1] apply sampling to propagate uncertainties in speech enhancement networks. Su *et al.* [48] use a truncated Gaussian graphical model in an expectation maximization framework. While elegant, the assumption of truncated variables appears difficult to be generalized to a wide range of nonlinear transformations such as max pooling, for instance.

Wang *et al.* [51] propose natural parameter networks (NPN), which treat the inputs, targets, weights, and neurons probabilistically by exponential-family distributions, *e.g.* assuming a Gaussian distribution for both weights and activations. In contrast to these assumptions, we aim for a lightweight approach without the Gaussian treatment of the parameters, which would increase the number of parameters significantly. Instead, we argue that probabilistic activations or even just a probabilistic output and loss suffice to reap some of the benefits of a probabilistic treatment.

By employing Gaussian activations, Jin *et al.* [27] show an improvement of classification robustness for adversarial inputs [17, 50]. While we also apply uncertainty propagating layers using Gaussians, there are some significant differences: Building upon standard maximum conditional likelihood learning [34], our work concentrates on probabilistic outputs. Combining it with uncertainty propagation throughout all layers is optional. Note that while in [27] un-certainties are dropped at the softmax layer, we provide an alternative classification layer that can be used in conjunction with maximum conditional likelihood learning.

## 3. Uncertainty Prediction in Supervised CNNs

We first describe the key ingredients that allow us to not only perform prediction with deep networks, but also assess predictive uncertainties. Thereby, we aim to alter existing networks as little as possible to ease adoption and aid practicality. We present two approaches: The first and simplest consists of solely replacing the output layer of well-proven networks with a probabilistic one. The second goes beyond this by considering activation uncertainties also *within* the network by means of deep uncertainty propagation.

### 3.1. CNNs with probabilistic outputs

Predictions from standard CNNs can be regarded as point estimates. While this is clearly true for regression networks, *e.g.* [10, 11], to some extent this also holds for classification networks with a softmax layer, *e.g.* [37, 45]. While the softmax yields class probabilities, it is mostly a convenient, smooth approximation to an $\arg\max$. The softmax predicts whether classes are more likely in relation to each other, but does not predict how certain the network is in that assessment. Consequently, softmax outputs are known not to be well calibrated [4, 13, 19].

**Notation.** Assume that we can formalize a deep neural network as a cascade of nonlinear layers

$$\mathbf{y} = \mathbf{f}(\mathbf{x}; \boldsymbol{\theta}) = \mathbf{f}^{(l)}\bigg(\mathbf{f}^{(l-1)}\Big(\cdots \mathbf{f}^{(1)}(\mathbf{x}; \boldsymbol{\theta}^{(1)})\Big)\bigg), \quad (1)$$

where $l$ is the number of layers, $\mathbf{x}$ are the inputs given to the network, and $\mathbf{y}$ are the predictions, which can be discrete or continuous. Each layer $\mathbf{f}^{(i)}(\mathbf{z}^{(i-1)}; \boldsymbol{\theta}^{(i)})$ corresponds to a nonlinear transformation of intermediate activations $\mathbf{z}^{(i-1)}$ (with $\mathbf{z}^{(0)} = \mathbf{x}$), possibly parameterized by $\boldsymbol{\theta}^{(i)}$. For brevity we summarize parameters of all layers with the vector $\boldsymbol{\theta}$.

**Probabilistic output layers.** Intuitively, one way to render Eq. (1) into an architecture with probabilistic outputs is to replace the point predictions $\mathbf{y}$ by probability distributions

$$\mathbf{f}(\mathbf{x}; \boldsymbol{\theta}) \equiv p(\cdot \,|\, \mathbf{x}), \quad (2)$$

which assign a probability (density) to all possible outputs $\mathbf{y}$. For tractability, we restrict the predictive distribution $p(\mathbf{y}\,|\,\mathbf{x})$ to be parametric, *i.e.* $p(\mathbf{y}\,|\,\mathbf{x}) \equiv p(\mathbf{y}\,|\,\mathbf{x}; \boldsymbol{\vartheta}_y)$, and let the last network layer $\mathbf{f}^{(l)}$ predict the parameters $\boldsymbol{\vartheta}_y$ instead. The parameters $\boldsymbol{\vartheta}_y$ encode a predicted output and its associated uncertainty. For instance, we may choose the parameters to be the (central) moments of a Gaussian with the variance corresponding to the uncertainty around the mean.

**Discussion.** Such a network still effectively predicts points in some sense. These points are not the outputs $\mathbf{y}$ directly, but the parameters $\boldsymbol{\vartheta}_y$ of a distribution over the outputs. While this is a simple change over standard deep networks, we show it to have significant practical benefits.

In practice, replacing deterministic outputs (Eq. 1) with a probabilistic output layer (Eq. 2) is rather straightforward once a parametric model is chosen. Importantly, the number of network parameters increases only in a minor way, in particular only by those from predicting $\boldsymbol{\vartheta}_y$ instead of $\mathbf{y}$ in the last layer $\mathbf{f}^{(l)}$. A Gaussian output layer, for instance, requires two minimal changes: *(i)* the number of outputs of the last layer has to be doubled accounting for both mean and variance, and *(ii)* since variances should be positive, we predict in $\log$ space, *i.e.* $\boldsymbol{\vartheta}_y = (\boldsymbol{\mu}, \widehat{\boldsymbol{v}})$ with $\boldsymbol{v} = \exp(\widehat{\boldsymbol{v}})$. The derivatives required for learning using SGD or variants are conveniently obtained by automatic differentiation.

### 3.2. Deep uncertainty propagation using ADF

While Eq. (2) accounts for uncertainty in the outputs, it does not allow for a probabilistic interpretation of intermediate activations, see Fig. 1. Intermediate activations $\mathbf{z}^{(i)}$ take a role in predicting the output activations, *e.g.* $\boldsymbol{\mu}$, their associated uncertainties, *e.g.* $\boldsymbol{v}$, or a mixture thereof. To put it differently, one cannot query intermediate layers on how certain they are about the presence of a feature, since this is encoded concurrently into their states. While this can be beneficial (we allow ultimate flexibility in modeling probabilistic outputs), a clearer separation of within-network activations and uncertainties may be desirable. Consequently, we apply the paradigm of replacing activations with distributions over activations to all layers. Similar to probabilistic outputs, each intermediate layer $\mathbf{f}^{(i)}, i = 1, \ldots, l-1$, then also outputs a distribution represented by some parameters $\boldsymbol{\vartheta}_{z^i}$ rather than a point estimate $\mathbf{z}^{(i)}$.

**Variational uncertainty propagation.** We now introduce our approach for propagating activation uncertainty in deep networks. To define a principled mechanism that can replace the standard activation propagation (Eq. 1) of the intermediate layers, we rely on a form of expectation propagation (EP) [39]. To keep the computation lightweight, we will restrict the EP framework to assumed density filtering (ADF) [7], which consists of a single EP forward pass. To formalize the deep probabilistic model, we start from a standard architecture that propagates point activations. Specifically, the joint density of all activations is given by

$$p(\mathbf{z}^{(0:l)}) = p(\mathbf{z}^{(0)}) \prod_{i=1}^{l} p(\mathbf{z}^{(i)}\,|\,\mathbf{z}^{(i-1)}), \quad (3a)$$

$$p(\mathbf{z}^{(i)}\,|\,\mathbf{z}^{(i-1)}) = \delta\Big[\mathbf{z}^{(i)} - \mathbf{f}^{(i)}(\mathbf{z}^{(i-1)})\Big], \quad (3b)$$

where $\delta[\cdot]$ denotes the Dirac delta. Note that inputs $p(\mathbf{z}^{(0)})$

in a deterministic network correspond to Dirac delta distributions. However, since inputs are never perfect, we assume them to be corrupted by white Gaussian noise

$$p(\mathbf{z}^{(0)}) = \prod_j \mathcal{N}\left(z_j^{(0)} \mid x_j, \sigma_n^2\right). \tag{4}$$

To propagate this aleatoric uncertainty through the network, we proceed to apply ADF to the network activations. The overall goal of the ADF framework is to find a tractable approximation of the network activations

$$p(\mathbf{z}^{(0:l)}) \approx q(\mathbf{z}^{(0:l)}) = q(\mathbf{z}^{(0)}) \prod_{i=1}^{l} q(\mathbf{z}^{(i)}), \tag{5}$$

incorporating one factor (or layer) of $p(\mathbf{z}^{(0:l)})$ at a time. Starting from the independent Gaussian input activations ($q(\mathbf{z}^{(0)}) = p(\mathbf{z}^{(0)})$), we approximate subsequent layer activations repeatedly by independent Gaussian distributions:

$$q(\mathbf{z}^{(i)}) = \prod_j \mathcal{N}\left(z_j^{(i)} \mid \mu_j^{(i)}, v_j^{(i)}\right), \tag{6}$$

where $\left(\mu_j^{(i)}, v_j^{(i)}\right)$ corresponds to the activation value and activation variance of neural unit $j$, respectively. For brevity we will summarize these independent activation distributions with vectors $\boldsymbol{\vartheta}_{\mathbf{z}^{(i)}} = \left(\boldsymbol{\mu}^{(i)}, \boldsymbol{v}^{(i)}\right)$ from now on.

The underlying process here is that subsequent layers $\mathbf{f}^{(i)}$ take an *activation distribution* $q(\mathbf{z}^{(i-1)})$ and transform it nonlinearly into an output distribution with a joint probability density $p(\mathbf{z}^{(i)} \mid \mathbf{z}^{(i-1)}) q(\mathbf{z}^{(i-1)})$ that can take complex forms and is not necessarily independent anymore. ADF assumes that previous factors in the variational distribution correspond to a reasonable approximation, *i.e.*

$$\tilde{p}(\mathbf{z}^{(0:i)}) = p(\mathbf{z}^{(i)} \mid \mathbf{z}^{(i-1)}) \prod_{j=0}^{i-1} q(\mathbf{z}^{(j)}). \tag{7}$$

Under this assumption, ADF then performs incremental updates of the variational approximation by solving

$$\underset{\tilde{q}(\mathbf{z}^{(0:i)})}{\arg \min} \quad \mathrm{KL}\left(\tilde{p}(\mathbf{z}^{(0:i)}) \parallel \tilde{q}(\mathbf{z}^{(0:i)})\right). \tag{8}$$

This layerwise approximation yields a canonical recipe enabling us to convert any layer with point activations (Eq. 1) into a layer that propagates uncertainties around their activations. Minka [39] has shown that the solution of Eq. (8) requires moment matching between $\tilde{p}(\mathbf{z}^{(0:i)})$ and $\tilde{q}(\mathbf{z}^{(0:i)})$. Assuming Eqs. (3a) and (6), this results in the following recipe: A layer $\mathbf{z}^{(i)} = \mathbf{f}^{(i)}(\mathbf{z}^{(i-1)}; \boldsymbol{\theta})$ can be converted into an uncertainty propagation layer by simply matching first and second-order central moments:

$$\boldsymbol{\mu}_z^{(i)} = \mathbb{E}_{q(\mathbf{z}^{(i-1)})}\left[\mathbf{f}^{(i)}(\mathbf{z}^{(i-1)}; \boldsymbol{\theta}^{(i)})\right] \tag{9a}$$

$$\boldsymbol{v}_z^{(i)} = \mathbb{V}_{q(\mathbf{z}^{(i-1)})}\left[\mathbf{f}^{(i)}(\mathbf{z}^{(i-1)}; \boldsymbol{\theta}^{(i)})\right], \tag{9b}$$

where $\mathbb{E}[\cdot]$ and $\mathbb{V}[\cdot]$ denote expectation and variance. In the final inference scheme, we pass through the neural network once, applying the local variational approximation from Eq. (8) at each subsequent layer. By doing so, ADF performs a greedy optimization of the global variational objective $\arg \min_{q(\mathbf{z}^{(0:l)})} \mathrm{KL}(p(\mathbf{z}^{(0:l)}) \parallel q(\mathbf{z}^{(0:l)}))$ [39].

Note that the true posterior of network activations can be multimodal. Here, the ADF framework approximates the density spanned by multiple modes [39], which contrasts variational inference methods that tend to be "mode greedy". While [14, 21] also apply ADF in a Bayesian context, they in contrast approximate the distribution of weights rather than the activation distributions. Our method only approximates activations, leading to a simple, yet effective method with a smaller computational footprint.

**Variational approximation of common layers.** Applying Eqs. (9a) and (9b) to create uncertainty propagation layers can often be done in closed form. Dense layers, convolutions, and deconvolutions are linear operations ($\mathbf{f}(\mathbf{z}) = W\mathbf{z} + \mathbf{b}$) and as such their moments are given by $\mathbb{E}[\mathbf{f}(\mathbf{z})] = W\boldsymbol{\mu}_z + \mathbf{b}$ and $\mathbb{V}[\mathbf{f}(\mathbf{z})] = (W \circ W)\boldsymbol{v}_z$, where $\circ$ is an elementwise product. Note that in such linear layers, the mean prediction and its variance do not interact. Other layers, *e.g.* pooling layers, do not necessarily have closed-form solutions, and adequate approximations for Eqs. (9a) and (9b) have to be found. For maxpool layers, a closed form solution exists for a two-element input, *i.e.* mean and variance of the max of two Gaussian distributed inputs can be analytically derived [25]. Jin *et al.* [27] generalize the two-input solution to more dimensions by folding the max operation across all elements inside the pooling region and applying the max operator in ascending order of the magnitude of the means. While such an ordering reduces the approximation error, we found this not to affect the resulting performance significantly. For this reason we simply fold the analytical solution first in horizontal and then in vertical direction to maximize the number of parallel operations.

ReLU nonlinearities $\mathrm{relu}(x) = \max(0, x)$ lead to closed form solutions, on the other hand. Frey and Hinton [12] showed that the moments of rectifiers under Gaussian activations with mean $\mu$ and variance $v$ are given by

$$\mu_{\mathrm{relu}}(\mu, v) = \mu \cdot \Phi\left(\frac{\mu}{\sigma}\right) + \sigma \cdot \phi\left(\frac{\mu}{\sigma}\right) \tag{10a}$$

$$v_{\mathrm{relu}}(\mu, v) = (\mu + v) \cdot \Phi\left(\frac{\mu}{\sigma}\right) + \mu\sigma \cdot \phi\left(\frac{\mu}{\sigma}\right) - \mu_{\mathrm{relu}}^2(\mu, v), \tag{10b}$$

where $\sigma = \sqrt{v}$ and $\phi(x), \Phi(x)$ are the standard normal and cumulative normal distribution, respectively. Note that in this nonlinear layer, mean and variance *do* interact.

**Leaky ReLU.** More recently, alternative nonlinear activation functions have emerged [20]. As an example, we derive the uncertainty propagation layer for the leaky ReLU:

$$\text{leaky\_relu}(x; c) = \max(c \cdot x, x), \quad 0 < c \ll 1. \quad (11)$$

Eq. (11) can be reformulated by means of common ReLUs:

$$\text{leaky\_relu}(x; c) = \text{relu}(x) - c\,\text{relu}(-x). \quad (12)$$

Equipped with Eqs. (10a) and (10b) we can show that

$$\mu_{\text{leaky\_relu}}(\mu, v) = \mu_{\text{relu}}(\mu, v) - c\,\mu_{\text{relu}}(-\mu, v) \quad (13a)$$
$$v_{\text{leaky\_relu}}(\mu, v) = v_{\text{relu}}(\mu, v) + c^2\,v_{\text{relu}}(-\mu, v)$$
$$+ 2c\,\mu_{\text{relu}}(\mu, v)\,\mu_{\text{relu}}(-\mu, v). \quad (13b)$$

In the supplemental material we give a comprehensive overview and some implementation details on the uncertainty propagation layers used in this work. Note that for numerical reasons we add a small constant (1e−4) to all activation uncertainties in all layers.

**Discussion.** In contrast to the simple probabilistic output layers from Sec. 3.1, uncertainty propagation layers based on ADF (Eqs. 9a, 9b) require *no additional parameters* (other than the input noise variance $\sigma_n^2$). This is because the last layer before the loss function already outputs a distribution by design. Moreover, *no changes* in the model architecture are required. Each layer can simply be replaced by its probabilistic counterpart in a drop-in fashion. All such layers are compatible by construction, *i.e.* they take in two values per neural unit and output two values.

## 4. Supervised Probabilistic Training

We now discuss how the probabilistic networks of Sec. 3 can be trained. We first note that it is always possible to obtain a point estimate at test time from the predictive distribution, *e.g.* by computing $\widehat{\mathbf{y}} = \arg\max_{\mathbf{y}} p(\mathbf{y} \mid \mathbf{x})$. Hence, if the point estimate $\widehat{\mathbf{y}}$ can be differentiated w.r.t. the parameters, we could apply standard loss-based training. However, we argue that a training objective solely based on a point estimate has disadvantages compared to a training objective using the probability (density) from Eqs. (2) and (3a).

**Maximum conditional likelihood learning.** Let $\mathbb{D} = \{\mathbf{x}^{(n)}, \mathbf{t}^{(n)}\}_{n=1}^{N}$ be a dataset with multivariate inputs $\mathbf{x}$ and targets $\mathbf{t}$. Assuming *i. i. d.* data, maximum conditional likelihood learning [34] finds the parameters by maximizing the conditional likelihood of the data under a predictive model

$$\boldsymbol{\theta}_{\text{MCLE}} = \arg\max_{\boldsymbol{\theta}} \prod_{n=1}^{N} p(\mathbf{t}^{(n)} \mid \mathbf{x}^{(n)}; \boldsymbol{\theta}). \quad (14)$$

For numerical reasons, learning is performed by minimizing the negative log-likelihood. To gain insight into the differences to standard loss-based training, we compare the conditional log-likelihood for a Gaussian output layer

$$\text{CLLH}(\boldsymbol{\theta} \mid \mathbb{D}) = \sum_{n=1}^{N} \sum_{j} \frac{\left(t_j^{(n)} - \mu_j^{(n)}\right)^2}{v_j^{(n)}} + \log v_j^{(n)} \quad (15)$$

to a sum of squared differences

$$\text{SSD}(\boldsymbol{\theta}; \mathbb{D}) = \sum_{n=1}^{N} \sum_{j} \left(t_j^{(n)} - y_j^{(n)}\right)^2. \quad (16)$$

Here, the conditional log-likelihood amounts to the squared error weighted by its predicted precision plus an additional term that ensures that the overall predicted variance stays low. Intuitively, a network predicting $\boldsymbol{\vartheta}_y = (\boldsymbol{\mu}, \boldsymbol{v})$ has two options to reduce the loss: First, it can improve its mean prediction, or, second, it can instead predict a high variance for outputs where it suspects the mean prediction to produce large errors. While such relationships between standard loss functions and conditional likelihoods are quite well known, they are used surprisingly rarely in practice. We here advocate the practical benefits of taking the probabilistic view. To that end, we discuss two concrete instantiations of this principle for regression and classification.

### 4.1. Regression with power exponential outputs

In regression problems in vision [10, 11], commonly a $L_p$-loss is applied, often $p = 1$ or $2$. We now aim to provide a probabilistic analog. To that end, we employ the *general power exponential distribution family* introduced by Gómez *et al.* [16], a multivariate probability density with $d$ dimensions (here, outputs per pixel) parametrized by three parameters $(\boldsymbol{\mu}, \boldsymbol{\Sigma}, k)$. When $k = 1/2$, the power exponential distribution equals a multivariate Laplacian, which we use when a per-pixel $L_2$-norm is desired. To keep inference tractable, we restrict $\boldsymbol{\Sigma}$ to be diagonal, *i.e.* $\boldsymbol{\beta} = \text{diag}\,\boldsymbol{\Sigma}$. The power exponential output layer is then given by

$$p(\mathbf{y} \mid \boldsymbol{\mu}, \boldsymbol{\beta}) \propto \prod_{j=1}^{d} \beta_j^{-\frac{1}{2}} \exp\left\{-\frac{1}{2}\left(\sum_{j=1}^{d} \frac{(y_j - \mu_j)^2}{\beta_j}\right)^k\right\}. \quad (17)$$

In practice, we minimize

$$-\log p(\mathbf{y} \mid \boldsymbol{\mu}, \boldsymbol{\beta}) \propto \sum_{j=1}^{d} \log \beta_j + \left(\sum_{j=1}^{d} \frac{(y_j - \mu_j)^2}{\beta_j}\right)^k. \quad (18)$$

Note that the per-pixel output dimensions in Eq. (18) are not independent, unlike in other recent work employing probabilistic outputs [42]. While other generalizations of a Laplacian to higher dimensions exist, we found the power exponential unit (with $k = 1/2$) to perform particularly well, *e.g.*, for the endpoint error in optical flow.

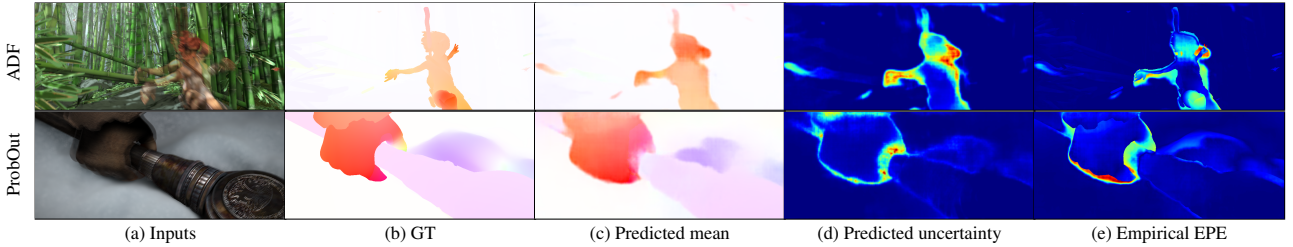

Figure 2. **Probabilistic regression of optical flow.** Our lightweight probabilistic CNNs, FlowNetADF and FlowNetProbOut, yield uncertainties for predictions while staying competitive w.r.t. the endpoint error (EPE). The uncertainties are highly correlated with the EPE.

### 4.2. Classification with Dirichlet outputs

In classification tasks a common output layer is the softmax, which normalizes point activations to obtain probabilities. The training loss is given by the average cross entropy between the predicted distribution and the empirical data distribution. Intuitively, the softmax and cross entropy allow to express some form of uncertainty: The network can produce low entropy, peaky distributions when it is confident; its predictions are driven towards high entropy, uniform distributions, when it is uncertain. However, probability distributions from a softmax are not well calibrated and high entropy distributions do not necessarily correlate with the actual error (*c.f.* Sec. 5.2). To address this issue, we use a continuous density defined on the probability simplex.

**Class uncertainties with Dirichlet distributions.** In this work, we use the Dirichlet distribution $\mathrm{Dir}(\mathbf{y}\,|\,\boldsymbol{\alpha})$ with concentration parameters $\alpha_j > 0$, a categorical distribution over the probability simplex $\sum_j y_j = 1, y_j > 0$. Note that for ADF it is not immediate how to propagate Gaussian activations ($n$ means and $n$ variances) to $n$ Dirichlet concentration parameters. Hence, inspired by [38], we reparametrize the Dirichlet distribution as

$$s = \left(\sum_j \alpha_j\right)^{-1}, \quad \mathbf{m} = s\,\boldsymbol{\alpha}, \quad s, m_j > 0, \quad (19)$$

where $\boldsymbol{\vartheta} = (\mathbf{m}, s)$ are $n$ location and a scale parameter, respectively. The location $\mathbf{m}$ with $\sum_j m_j = 1$ corresponds to a point prediction on the probability simplex, and $s$ is a scalar "second order central moment" indicating how stretched or squeezed the density is around $\mathbf{m}$.

**Dirichlet outputs.** To only require minimal changes to existing networks, we avoid predicting the scale $s$ directly. Instead, we equip each neuron with its individual uncertainty as before. To aggregate the $n$ uncertainties into the required scale $s$, we define a *variance pooling* mechanism. We formalize the Dirichlet output layer as

$$p(\cdot\,|\,\mathbf{z}) = \mathrm{Dir}(\cdot\,|\,\boldsymbol{\alpha}(\boldsymbol{\mu}_z, \boldsymbol{v}_z)) \text{ with } \boldsymbol{\alpha}(\boldsymbol{\mu}_z, \boldsymbol{v}_z) = \frac{\mathbf{m}}{s} \quad (20)$$

and

$$\mathbf{m} = \mathrm{softmax}(\boldsymbol{\mu}_z), \quad s = c_1 + c_2 \sqrt{\sum_j m_j\, v_j}, \quad (21)$$

where $c_1 > 0$ controls how peaky the resulting distributions on the simplex can become and $c_2 > 0$ is an amplification factor for converting uncertainties into the Dirichlet scale. Note that the Dirichlet scale $s$ is computed from the weighted uncertainty of individual neurons; hence the influence of an individual uncertainty depends on its softmax activation. Thus, the variance we care most about is the one contributing most to the softmax activation.

**Details.** We may now directly predict $\boldsymbol{\vartheta}_z = (\boldsymbol{\mu}_z, \boldsymbol{v}_z)$ and perform maximum conditional likelihood learning, *i.e.* maximizing Eq. (14) under the Dirichlet distribution. The Dirichlet distribution only supports continuous vectors on the unit simplex; hence we render the discrete ground truth labels into continuous multinomial parameters by applying Laplace smoothing [26] with a small $\delta$ (1e−3) to the labels. The smoothing parameter requires a trade-off between robustness and accuracy and is chosen such that networks are as robust as possible, while still performing competitively. The training objective is then the maximum conditional likelihood of these 'ground truth' multinomial parameters under the predictive Dirichlet distribution.

**Discussion.** Note that while it may seem natural to apply a conjugate Dirichlet-Multinomial instead of a Dirichlet with smoothed labels here, this is not an option. The issue is that for a single trial this would fall back to a softmax and cross entropy. On the other hand, a network with the proposed Dirichlet output can decide to output a fairly peaky *point prediction* $\mathbf{m}$ when it is certain that the resulting class is likely a specific one, yet at the same time output a high scale parameter $s$. One may wonder when this may make sense. An answer lies in the fact that the network can only pick classes from a restricted set that it knows from training [4]. Hence the network may predict that one class is more likely than the others, but that it is overall not that confident.

## 5. Experiments

### 5.1. Probabilistic regression with FlowNet

To demonstrate the benefits of the proposed probabilistic framework for a regression task, we conduct experiments with optical flow. While various energy-based meth-ods have been proposed that typically exploit the brightness constancy assumption [49], Dosovitskiy *et al.* [10] introduced FlowNet, a CNN architecture that learns to predict optical flow from data directly. The CNN is discriminatively trained on the synthetic FlyingChairs dataset using the endpoint error (EPE) as loss. We take their *FlowNetS* architecture and convert it into two different probabilistic versions: The first architecture, *FlowNetADF*, consists of translating each layer into its uncertainty propagation counterpart (*c.f.* Sec. 3.2), where we set the input noise to $\sigma_n = 0.01$. Note that the input noise parameter is not too sensitive as long as it is small in relation to the inputs. In our second architecture, termed *FlowNetProbOut*, we do not propagate uncertainties, but apply probabilistic output layers alone (*c.f.* Sec. 3.1). For both networks, we replace the endpoint error (of the standard FlowNetS) by a Laplacian output layer. Here, we use the power exponential layer (*c.f.* Sec. 4.1) with $k = 1/2$.

Table 1. **Endpoint error of optical flow regression using several variants of FlowNet.** "fps" denotes the test-time speed (GTX 1080 Ti) in frames per second with batches of size one.

| Network | fps | Sintel clean train | Sintel clean test | Sintel final train | Sintel final test | Chairs test |
|---|---|---|---|---|---|---|
| FlowNetS [10] | – | (4.50) | (7.42) | (5.45) | (8.42) | (2.71) |
| FlowNetS (PT) | **106** | 4.58 | 7.66 | 5.72 | 8.53 | 2.38 |
| FlowNetADF (ours) | 38 | **4.39** | **7.46** | 5.69 | 8.53 | 2.19 |
| FlowNetProbOut (ours) | 101 | 4.52 | 7.47 | **5.58** | **8.30** | **2.15** |
| FlowNetDropOut | 3 | 4.56 | 7.65 | 5.70 | 8.49 | 2.39 |

Table 2. **Avg. log likelihoods of probabilistic flow regression.**

| Network | Sintel clean train | Sintel final train | Chairs test |
|---|---|---|---|
| FlowNetADF (ours) | **−3.878** | **−4.186** | **−3.348** |
| FlowNetProbOut (ours) | −6.888 | −7.621 | −3.591 |
| FlowNetDropOut | −7.106 | −10.820 | −6.176 |

As a probabilistic baseline, we implemented a version with variational Gaussian dropout [31] at the bottleneck (*FlowNetDropOut*). Note that additional dropout layers resulted in a decrease in accuracy. We perform sampling (with 30 samples) and use their average as mean prediction. To obtain a predictive density, we perform kernel density estimation with a Gaussian kernel, where the bandwidth has been optimized w.r.t. the FlyingChairs validation set.

**Implementation.** As gradients of our ADF network are quite involved, we rely on automatic differentiation and reimplemented FlowNetS in PyTorch (*FlowNetS (PT)*). Our probabilistic networks and our reimplementation are trained with similar parameters (216 epochs of Adam [29]) on the FlyingChairs dataset, which allows a fair comparison.

**Results.** Fig. 2 shows how the probabilistic networks perform on select images from the Sintel [8] dataset. The two rightmost columns visualize the differential entropy of the predicted distributions and the endpoint error, respectively. Our predicted uncertainties are highly correlated with the actual endpoint error, which suggests that our probabilistic approach is able to assess where it fails and where it succeeds, *c.f.* supplemental for further results. Table 1 shows quantitative results for the Sintel datset and for a hold-out set from FlyingChairs (640 images); predictions of probabilistic networks are given by the means. We report the execution speed on a Nvidia GTX 1080 Ti.

FlowNetS (PT) yields better results on FlyingChairs than the original, but slightly worse results on Sintel likely due to minor differences in training. The probabilistic baseline FlowNetDropOut performs slightly better than FlowNetS (PT), but is an order of magnitude slower than our probabilistic networks. More importantly, our probabilistic networks perform competitive w.r.t. our reimplementation of FlowNetS; notably FlowNetADF outperforms the deterministic counterpart with a 4% improvement on Sintel clean. Hence, the predictive power of our probabilistic networks is as good or better than the corresponding deterministic architecture, making them attractive as a drop-in replacement. Their benefit is that they allow to assess the uncertainty of their prediction, *i.e.* they know when and where they fail. Table 2 shows the average log likelihoods of the probabilistic networks on Sintel and FlyingChairs, where FlowNetADF clearly outperforms both FlowNetProbOut and FlowNetDropOut. Our lightweight probabilistic networks are highly practical and do not add much to the execution time (*c.f.* frames per second in Table 1) in contrast to other approaches, such as test-time dropout [13]. The speed difference between FlowNetProbOut and FlowNetS (PT) is minimal. The uncertainty propagation of FlowNetADF renders its speed to roughly a third due to doubling the number of activations in every layer with additional nonlinearities.

### 5.2. Probabilistic classification

We perform classification experiments on CIFAR10 and MNIST. For MNIST we use the LeNet architecture [36], while we use a fully convolutional architecture (All-CNN-C) [47] for CIFAR10. We compare 4 different architectures: *(Determ.)* The deterministic baseline network; *(ProbOut)* our probabilistic model where the layer before the softmax is a Gaussian layer as in Sec. 3.1; *(ADF)* the uncertainty propagation network from Sec. 3.2 (with $\sigma_n = 0.01$); and *(Dropout)* test-time dropout [13] with $p = 0.5$. Determ. and Dropout are trained via the standard cross-entropy (XE). For the probabilistic networks, we apply 3 different training paradigms: *(SM mean + XE)* We directly feed the mean-prediction into a softmax as suggested by [27] and use the cross-entropy loss; *(SM approx + XE)* we apply the second-order softmax approximation of [44] with a cross-entropy loss; *(Dir + CLLH)* our Dirichlet output layer from Sec. 4.2 trained by conditional likelihood maximization. Despite using a different loss for the Dirichlet output layer, we can compare the classification accuracy and the cross-entropy of the class predictive distributions.

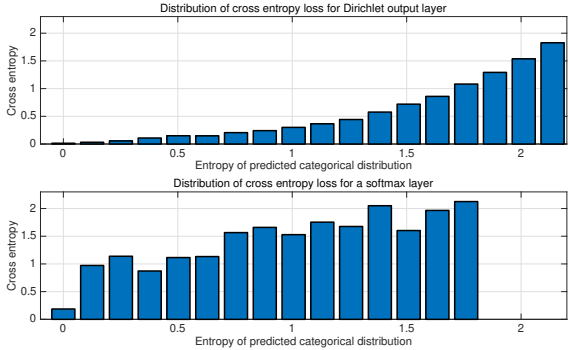

Figure 3. **Assessment of predictive classification distributions on CIFAR10:** Entropy of the categorical distribution (x-axis) *vs.* cross-entropy of the data (y-axis). While the softmax (bottom) shows a weak correlation between its uncertainty and the empirical error, the Dirichlet output layer (top) shows a strong correlation.

**Results.** Table 3 shows quantitative results on CIFAR10 and MNIST. While classification performance is similar for networks trained with the same loss, Dir + CLLH outperforms the softmax based losses. Interestingly, a lower cross entropy does not always yield a better classification accuracy. Test-time dropout performs slightly better in terms of cross entropy on CIFAR, but is not very practical, as it requires many, often hundreds forward passes through the network. As for regression, we find that our lightweight probabilistic networks retain the predictive power of their deterministic counterpart. To analyze the predicted distributions, we measure the entropy of the networks' prediction *vs.* the actual error. Fig. 3 shows the cross-entropy loss *vs.* the entropy of the predictive categorical distribution for the test data of CIFAR10, for both the Dirichlet output layer (ADF) well as a softmax layer (Determ.). Intuitively, the cross-entropy loss should be positively correlated with the entropy of the predictive categorical distribution, *i.e.* the network outputs a high entropy for a large empirical error. While for the Dirichlet output layer the empirical error is clearly correlated with the predictive entropy, only a weak correlation can be observed for the softmax. Hence, while the softmax yields high accuracy predictions, it does not calibrate well for the actual uncertainties contained in the prediction. Since our Dirichlet output layer is more flexible in modeling the predictive distribution, it allows for a better calibration, while still achieving similar or better accuracy.

**Adversarial GSM attack.** To evaluate the robustness of networks against adversarial attacks, we apply the gradient sign method (GSM) of [17] to the architectures from above. Table 4 shows the classification performance for various networks and three different attack strengths $\epsilon$. We observe that the Dirichlet layer is more robust against this attack than the softmax layer, which highlights the strength of our lightweight probabilistic approach. We finally performed an attack to the test-time dropout architecture by backpropagating the gradients through the samples. While test-time dropout yields high accuracy and low cross-entropy, it appears more prone to adversarial attacks than Dir + CLLH.

Table 3. **Classification accuracy** (in %) **and cross-entropy** (XE).

| | | CIFAR10 test | | MNIST test | |
|---|---|---|---|---|---|
| Network | Output / Loss | Acc. | XE | Acc. | XE |
| Determ. | SM + XE | 90.62 | 0.369 | 99.39 | 0.0229 |
| Dropout | SM + XE | 90.88 | **0.327** | 99.40 | 0.0222 |
| ADF | SM approx + XE | 89.25 | 0.431 | 99.41 | 0.0263 |
| ProbOut | SM approx + XE | 89.62 | 0.460 | 99.22 | 0.0223 |
| ADF | SM mean + XE | 89.16 | 0.467 | 99.30 | 0.0241 |
| ADF | Dir + CLLH (ours) | 91.51 | 0.450 | **99.50** | 0.0247 |
| ProbOut | Dir + CLLH (ours) | **91.87** | 0.366 | 99.48 | **0.0202** |

Table 4. **Adversarial FGSM attack on CIFAR10.** Accuracy (in %) after an attack using the gradient sign method from [17].

| | | $\epsilon$ | | |
|---|---|---|---|---|
| Network | Output / Loss | 0.1 | 0.05 | 0.01 |
| Determ. | SM + XE | 23.99 | 35.79 | 46.66 |
| Dropout | SM + XE | 22.48 | 31.08 | 42.24 |
| ADF | SM approx + XE | 5.04 | 4.48 | 19.10 |
| ProbOut | SM approx + XE | 4.90 | 4.60 | 22.47 |
| ADF | SM mean + XE | 6.05 | 6.74 | 20.63 |
| ADF | Dir + CLLH (ours) | **25.20** | 43.03 | 55.91 |
| ProbOut | Dir + CLLH (ours) | 23.20 | **46.26** | **66.49** |

## 6. Conclusion

In this paper we have proposed a lightweight treatment of probabilities in supervised deep networks. We suggested probabilistic output layers and introduced a deep uncertainty propagation procedure derived from assumed density filtering. We showed how regression problems can be formulated with a general power exponential layer, and classification tasks can be approached with a Dirichlet layer. Both can be trained by conditional likelihood maximization. We demonstrated how these models can be used as efficient drop-in replacements for deep networks in vision. They retain the predictive power and (most of the) computational efficiency of the original architecture, but allow for assessing the uncertainty of the prediction, which is highly correlated to the empirical error. Classification networks additionally become more robust against adversarial attacks.

**Acknowledgments.** The research leading to these results has received funding from the European Research Council under the European Union's Seventh Framework Programme (FP/2007–2013)/ERC Grant agreement No. 307942.

# Lightweight Probabilistic Deep Networks
## – Supplemental Material –

Jochen Gast      Stefan Roth

Department of Computer Science, TU Darmstadt

In this supplemental material we derive the recipe to create uncertainty propagating layers, give additional details on the various types of layers used in the main paper, visualize the effect of uncertainty prediction, and illustrate its benefits with example classification and regression results.

## A. Assumed Density Filtering

We now derive the necessary equations to show that our framework indeed corresponds to ADF applied to the joint density of neural network activations. Let us briefly recall the assumptions of our main paper. The true joint density of network activations in a deterministic network with Gaussian independent inputs is given by

$$p(\mathbf{z}^{(0:l)}) = p(\mathbf{z}^{(0)}) \prod_{i=1}^{l} p(\mathbf{z}^{(i)} \mid \mathbf{z}^{(i-1)}) \tag{22a}$$

with

$$p(\mathbf{z}^{(i)} \mid \mathbf{z}^{(i-1)}) = \delta\big[\mathbf{z}^{(i)} - \mathbf{f}^{(i)}(\mathbf{z}^{(i-1)})\big] \tag{22b}$$

$$p(\mathbf{z}^{(0)}) = \prod_j \mathcal{N}\big(z_j^{(0)} \mid x_j, \sigma_n^2\big). \tag{22c}$$

The approximate joint density is assumed to be

$$q(\mathbf{z}^{(0:l)}) = q(\mathbf{z}^{(0)}) \prod_{i=1}^{l} q(\mathbf{z}^{(i)}), \qquad q(\mathbf{z}^{(i)}) = \prod_j \mathcal{N}\big(z_j^{(i)} \mid \mu_j^{(i)}, v_j^{(i)}\big). \tag{23}$$

As explained in the main paper, we initialize the first factor of our approximation by setting $q(\mathbf{z}^{(0)}) = p(\mathbf{z}^{(0)})$. This is possible since we assume Gaussian input noise. Afterwards, ADF performs iterative approximations by minimizing

$$\underset{\tilde{q}(\mathbf{z}^{(0:i)})}{\arg\min} \quad \text{KL}(\tilde{p}(\mathbf{z}^{(0:i)}) \,\|\, \tilde{q}(\mathbf{z}^{(0:i)})), \tag{24}$$

where the true posterior at each iteration consists of the newest factor and the previous approximating factors

$$\tilde{p}(\mathbf{z}^{(0:i)}) = p(\mathbf{z}^{(i)} \mid \mathbf{z}^{(i-1)}) \prod_{j=0}^{i-1} q(\mathbf{z}^{(j)}). \tag{25}$$

Updates are then performed for the consecutive iterations (or layers) $i = 1, \ldots, l$. For the Gaussian assumption (Eq. 23), the solution to Eq. (24) equals to matching the first-order and second-order moments of $\tilde{p}(\mathbf{z}^{(0:i)})$ and $\tilde{q}(\mathbf{z}^{(0:i)})$.

We now show that for neural network activations as modeled by Eqs. (22a) to (22c), solving Eq. (24) exactly yields our recipe from Eqs. (9a) and (9b) of the main paper for creating uncertainty propagating layers. To show this, the following two properties of Dirac delta distributions are quite useful:

$$\int_{-\infty}^{\infty} \delta[x-y] g(x) \,\mathrm{d}x = g(y) \quad \text{and} \quad \int_{-\infty}^{\infty} \delta[x-y] \,\mathrm{d}x = 1. \tag{26}$$

i

Also note that we will factorize the approximate posterior by removing an activation variable $z_k$ from the distribution and explicitly represent it as a separate factor, *i.e.* $q(\mathbf{z}) = q(\mathbf{z}_{\searrow k}) q(z_k)$, where $q(\mathbf{z}_{\searrow k})$ corresponds to the joint density of all factors excluding $z_k$.

We continue to derive the first order moment of $\tilde{p}(\mathbf{z}^{(0:i)})$, which will be done in two steps. First, we derive the moment of activation variables $z_k$ of all layers excluding the last layer, *i.e.* for $z_k$ that are an element of $\mathbf{z}^{(0:i-1)}$. Afterwards, we take a look at activation variables solely contained in the last layer, *i.e.* for $z_k$ elements of $\mathbf{z}^{(i)}$.

For activation variables $z_k$ not contained in the last layer (*i.e.* part of $\mathbf{z}^{(0:i-1)}$), we have the first moment

$$\mathbb{E}_{\tilde{p}}[z_k] = \int_{\mathbf{Z}} \tilde{p}(\mathbf{z}^{(0:i)}) z_k \, d\mathbf{z} = \int_{-\infty}^{\infty} \int_{\mathbf{Z}_{\searrow k}} \tilde{p}(\mathbf{z}^{(0:i)}) z_k \, d\mathbf{z}_{\searrow k} \, dz_k \tag{27a}$$

$$= \int_{-\infty}^{\infty} \int_{\mathbf{Z}_{\searrow k}^{(0:i-1)}} \int_{\mathbf{Z}^{(i)}} \left( \delta[\mathbf{z}^{(i)} - \mathbf{f}^{(i)}(\mathbf{z}^{(i-1)})] q(z_k) q(\mathbf{z}_{\searrow k}^{(0:i-1)}) z_k \right) d\mathbf{z}^{(i)} \, d\mathbf{z}_{\searrow k}^{(0:i-1)} \, dz_k \tag{27b}$$

$$= \int_{-\infty}^{\infty} q(z_k) z_k \left( \int_{\mathbf{Z}_{\searrow k}^{(0:i-1)}} q(\mathbf{z}_{\searrow k}^{(0:i-1)}) \left( \int_{\mathbf{Z}^{(i)}} \delta[\mathbf{z}^{(i)} - \mathbf{f}^{(i)}(\mathbf{z}^{(i-1)})] \, d\mathbf{z}^{(i)} \right) d\mathbf{z}_{\searrow k}^{(0:i-1)} \right) dz_k \tag{27c}$$

$$= \int_{-\infty}^{\infty} q(z_k) z_k \left( \int_{\mathbf{Z}_{\searrow k}^{(0:i-1)}} q(\mathbf{z}_{\searrow k}^{(0:i-1)}) \, d\mathbf{z}_{\searrow k}^{(0:i-1)} \right) dz_k = \int_{-\infty}^{\infty} q(z_k) z_k \, dz_k = \mathbb{E}_{q(z_k)}[z_k]. \tag{27d}$$

Hence, for all layers except for the $i^{\text{th}}$ layer, the first moments remain unchanged after the update. For activation variables $z_k$ solely contained in the last layer $\mathbf{z}^{(i)}$, we have the first moment

$$\mathbb{E}_{\tilde{p}}[z_k] = \int_{\mathbf{Z}} \tilde{p}(\mathbf{z}^{(0:i)}) z_k \, d\mathbf{z} = \int_{\mathbf{Z}_{\searrow k}} \int_{-\infty}^{\infty} \tilde{p}(\mathbf{z}^{(0:i)}) z_k \, dz_k \, d\mathbf{z}_{\searrow k} \tag{28a}$$

$$= \int_{\mathbf{Z}^{(0:i-1)}} \int_{\mathbf{Z}_{\searrow k}^{(i)}} \int_{-\infty}^{\infty} \left( \delta[\mathbf{z}^{(i)} - \mathbf{f}^{(i)}(\mathbf{z}^{(i-1)})] q(\mathbf{z}^{(0:i-1)}) z_k \right) dz_k \, d\mathbf{z}_{\searrow k}^{(i)} \, d\mathbf{z}^{(0:i-1)} \tag{28b}$$

$$= \int_{\mathbf{Z}^{(0:i-1)}} q(\mathbf{z}^{(0:i-1)}) \left( \int_{\mathbf{Z}_{\searrow k}^{(i)}} \int_{-\infty}^{\infty} \left( \delta[\mathbf{z}^{(i)} - \mathbf{f}^{(i)}(\mathbf{z}^{(i-1)})] z_k \right) dz_k \, d\mathbf{z}_{\searrow k}^{(i)} \right) d\mathbf{z}^{(0:i-1)} \tag{28c}$$

$$= \int_{\mathbf{Z}^{(0:i-1)}} q(\mathbf{z}^{(0:i-1)}) \left( \int_{\mathbf{Z}_{\searrow k}^{(i)}} \int_{-\infty}^{\infty} \left( \delta[z_k - f_k^{(i)}(\mathbf{z}^{(i-1)})] \delta[\mathbf{z}_{\searrow k}^{(i)} - \mathbf{f}_{\searrow k}^{(i)}(\mathbf{z}^{(i-1)})] z_k \right) dz_k \, d\mathbf{z}_{\searrow k}^{(i)} \right) d\mathbf{z}^{(0:i-1)} \tag{28d}$$

$$= \int_{\mathbf{Z}^{(0:i-1)}} q(\mathbf{z}^{(0:i-1)}) \left( \int_{-\infty}^{\infty} \delta[z_k - f_k^{(i)}(\mathbf{z}^{(i-1)})] z_k \, dz_k \right) \left( \int_{\mathbf{Z}_{\searrow k}^{(i)}} \delta[\mathbf{z}_{\searrow k}^{(i)} - \mathbf{f}_{\searrow k}^{(i)}(\mathbf{z}^{(i-1)})] \, d\mathbf{z}_{\searrow k}^{(i)} \right) d\mathbf{z}^{(0:i-1)} \tag{28e}$$

$$= \int_{\mathbf{Z}^{(0:i-1)}} q(\mathbf{z}^{(0:i-1)}) f_k^{(i)}(\mathbf{z}^{(i-1)}) \, d\mathbf{z}^{(0:i-1)} = \int_{\mathbf{Z}^{(i-1)}} q(\mathbf{z}^{(i-1)}) f_k^{(i)}(\mathbf{z}^{(i-1)}) \, d\mathbf{z}^{(i-1)} = \mathbb{E}_{q^{(i-1)}}[f_k^{(i)}(\mathbf{z}^{(i-1)})], \tag{28f}$$

which exactly corresponds to our recipe in Eq. (9a). Note that by replacing $z_k$ with $z_k^2$ in the derivations above, it is easy to obtain similar results for the second order moments:

$$\forall z_k \text{ elements of } \mathbf{z}^{(0:i-1)}: \quad \mathbb{E}_{\tilde{p}}[z_k^2] = \mathbb{E}_{q(z_k)}[z_k^2] \tag{29a}$$

$$\forall z_k \text{ elements of } \mathbf{z}^{(i)}: \quad \mathbb{E}_{\tilde{p}}[z_k^2] = \mathbb{E}_{q^{(i-1)}}[f_k^2(\mathbf{z}^{(i-1)})], \tag{29b}$$

which yields the variance update of the recipe Eq. (9b).

## B. Uncertainty Propagation Layers

We now describe the details of various layer implementations. Recall the general recipe as given in the main paper, which we just derived as

$$\boldsymbol{\mu}_z^{(i)} = \mathbb{E}_{q(\mathbf{z}^{(i-1)})} \left[ \mathbf{f}^{(i)}(\mathbf{z}^{(i-1)}; \boldsymbol{\theta}^{(i)}) \right] \tag{30a}$$

$$\mathbf{v}_z^{(i)} = \mathbb{V}_{q(\mathbf{z}^{(i-1)})} \left[ \mathbf{f}^{(i)}(\mathbf{z}^{(i-1)}; \boldsymbol{\theta}^{(i)}) \right]. \tag{30b}$$

**Linear layers.** As explained in the main paper, fully connected layers, convolutions, and deconvolutions can all be formalized as linear functions

$$\mathbf{z}^{(i)} = W\mathbf{z}^{(i-1)} + \mathbf{b}. \tag{31}$$

By the linearity of the expectation, the uncertainty propagation layer is consequently given by

$$\boldsymbol{\mu}_z^{(i)} = W\boldsymbol{\mu}_z^{(i-1)} + \mathbf{b} \tag{32a}$$

$$\mathbf{v}_z^{(i)} = (W \circ W)\mathbf{v}_z^{(i-1)}. \tag{32b}$$

**(Global) average pooling.** The uncertainty propagation layer for average pooling

$$z^{(i)} = \mathrm{mean}\left(\mathbf{z}^{(i-1)}\right) \tag{33}$$

can be derived by formalizing it as a linear layer, where weights are chosen such that they correspond to the averaging operator. In practice, we can implement the layer via the following equations:

$$\mu_z^{(i)} = \mathrm{mean}(\boldsymbol{\mu}_z^{(i-1)}) \tag{34a}$$

$$v_z^{(i)} = \frac{1}{n}\mathrm{mean}(\mathbf{v}_z^{(i-1)}), \tag{34b}$$

where for Eq. (34b) we rely on the fact that the mean operator itself already multiplies $1/n$ to the variances. Here, $n$ is the number of elements of $\mathbf{z}^{(i-1)}$. Also note that $n$ differs w.r.t. different pooling sizes.

**Max pooling.** A max pooling unit can be understood as returning the maximum response of a number of inputs, *i.e.* for two inputs $A$ and $B$ we have

$$C = \max(A, B). \tag{35}$$

For $A \sim \mathcal{N}(\mu_A, \sigma_A^2)$ and $B \sim \mathcal{N}(\mu_B, \sigma_B^2)$ the probability density of their maximum $C$ according to [25] is given by

$$p(C = c) = \mathcal{N}(c \mid \mu_A, \sigma_A^2) \cdot \Phi\left(\frac{c - \mu_B}{\sigma_B}\right) + \mathcal{N}(c \mid \mu_B, \sigma_B^2) \cdot \Phi\left(\frac{c - \mu_A}{\sigma_A}\right), \tag{36a}$$

where

$$\Phi(y) = \int_{-\infty}^{y} \phi(x)\,\mathrm{d}x \quad \text{and} \quad \phi(y) = \frac{1}{\sqrt{2\pi}} \exp(-y^2/2) \tag{36b}$$

are the CDF and PDF of a unit-normal distribution.

Despite $C$ not being normally distributed anymore, we approximate it by a univariate normal. In [25] the first and second moment are derived analytically. The mean is given by

$$\mu_C = \sqrt{\sigma_A^2 + \sigma_B^2} \cdot \phi(\alpha) + (\mu_A - \mu_B) \cdot \Phi(\alpha) + \mu_B \quad \text{with} \quad \alpha = \frac{\mu_A - \mu_B}{\sqrt{\sigma_A^2 + \sigma_B^2}} \tag{37a}$$

and the variance can be written as

$$v_C = (\mu_A + \mu_B)\sqrt{\sigma_A^2 + \sigma_B^2} \cdot \phi(\alpha) + (\mu_A^2 + \sigma_A^2) \cdot \Phi(\alpha) + (\mu_B^2 + \sigma_B^2) \cdot \left(1 - \Phi(\alpha)\right) - \mu_C^2. \tag{37b}$$

If we have more than two inputs, we concatenate the operations. More precisely, we fold any max pooling operation first in horizontal and then in vertical direction.

**Rectified linear unit (ReLU).** *c.f.* main paper.

**Leaky rectifier (LeakyReLU).** *c.f.* main paper.

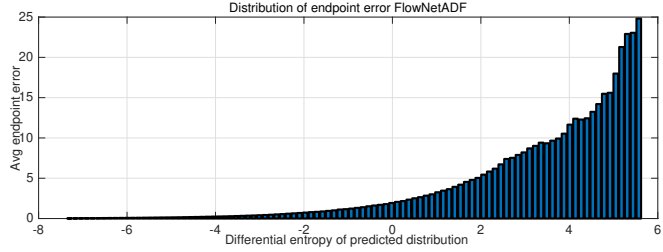

Figure 5. **Assessment of predicted regression uncertainties for an application in optical flow:** Differential entropy of the predictive distribution (x-axis) *vs.* average endpoint error (y-axis). A high entropy corresponds to a prediction with high variance. In this region the endpoint error is also large, suggesting that the uncertainty of the prediction is highly correlated to the actual error.

## C. Calibration of Regression Uncertainties

Similar to the comparison of cross entropy *vs.* categorical entropy for classification in the main paper, we can assess the quality of the predictive distributions of our probabilistic optical flow networks in the case of regression. As already pointed out in the main paper, our predicted uncertainties are well correlated with the actual endpoint error, which suggests that our probabilistic approach is able to assess where it fails and where it succeeds. This is also borne out in Fig. 5, where we plot the endpoint error for FlowNetADF against the differential entropy for all images in the FlyingChairs test set. Hence, predicted variances can be reliably used to assess a model's accuracy.

## D. Exponential Power Outputs

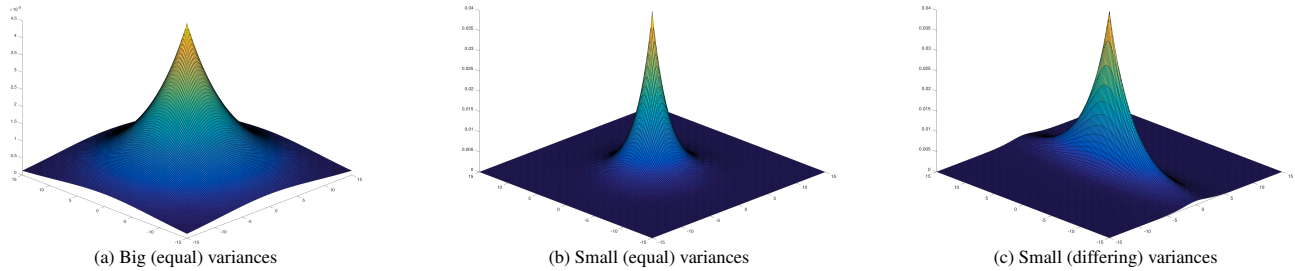

(a) Big (equal) variances      (b) Small (equal) variances      (c) Small (differing) variances

Figure 6. **Exponential power units for optical flow regression** in *FlowNetADF* and *FlowNetProbOut*. In this (artificial) illustrative example, the mean prediction is zero. (a) and (b) correspond to high and low variance predictions in which the variances of horizontal and vertical flow are equal, respectively. (c) shows the density for a case, where horizontal and vertical variances differ by an order of magnitude.

To gain some intuition on the workings of the probabilistic prediction, Fig. 6 shows a visualization of the probability density of the predictive distribution at an output node for the two-dimensional exponential power unit we used in the FlowNet networks of the main paper. The $x$-$y$ axes correspond to $u$-$v$ components of the optical flow. Fig. 6a shows the density of an uncertain prediction with large variance. Fig. 6b instead shows a (more) certain prediction with a smaller variance. In both cases, the variances for the horizontal and vertical flow directions are equal, respectively. Fig. 6c also corresponds to a smaller variance, however, here variances in horizontal and vertical direction differ by an order of magnitude. Note that unlike in a Gaussian output, the flow components in horizontal and vertical direction are not independent.

## E. Additional Examples

Figs. 7 and 8 shows some examples from the CIFAR10 test set to illustrate how softmax predictions differ from predictions made with our lightweight probabilistic networks. Note that for our networks we extract the categorical distribution from the mean of the Dirichlet. While softmax predictions tend to be very confident in most cases, *i.e.* predictions have very low entropy with much of the probability mass concentrated on a single category, predictions made with the Dirichlet output layer rarely collapse to a single class in practice. Also, the predictions tend to have higher entropy, *i.e.* a more uniform distribution over classes, when the network makes an incorrect prediction. In other words our lightweight probabilistic networks have a sense of when they fail.

Figs. 11 and 12 give additional results for optical flow prediction on the Sintel dataset.

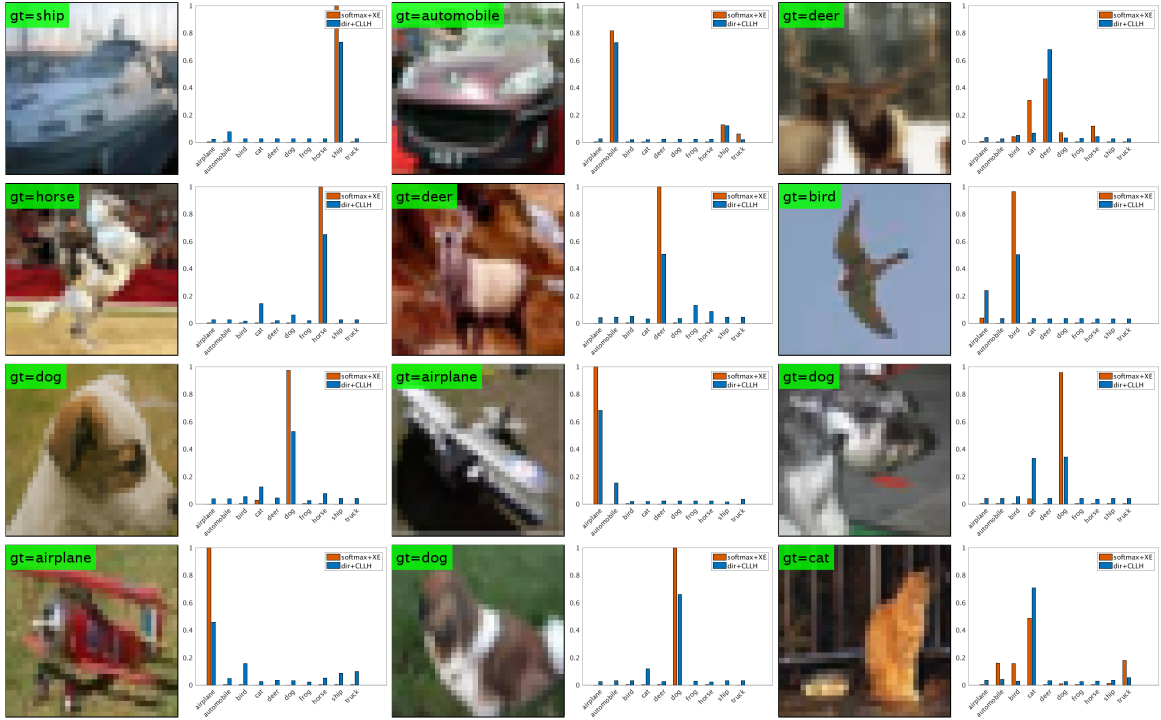

Figure 7. **Correctly classified CIFAR10 images** (test set). Blue bars indicate results from our ADF network with Dirichlet outputs (Dir + CLLH), while red bars indicate deterministic results with a softmax. Overall, the softmax results tend to be much more confident than the Dirichlet output layer, *i.e.* they often yield peaky predictions with very low entropy. While the Dirichlet output layer also assigns the correct Top-1 class, its predictions do not tend to be quite as confident as the softmax predictions, *i.e.* yielding higher entropy predictions.

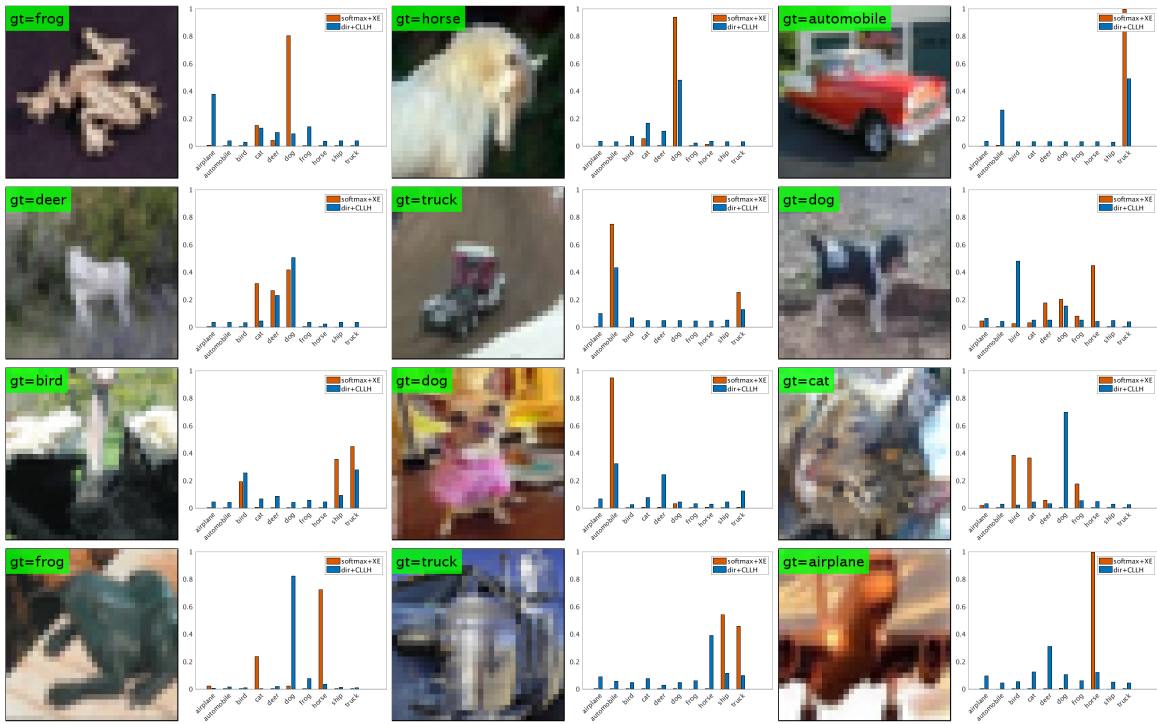

Figure 8. **Missclassified CIFAR10 images** (test set). Color coding as above. Although the predictions are incorrect, the deterministic softmax predictions are still highly confident (cf. frog top left, airplane bottom right). The Dirichlet layer also fails on these cases, yet its predictions are less confident and have higher entropy.

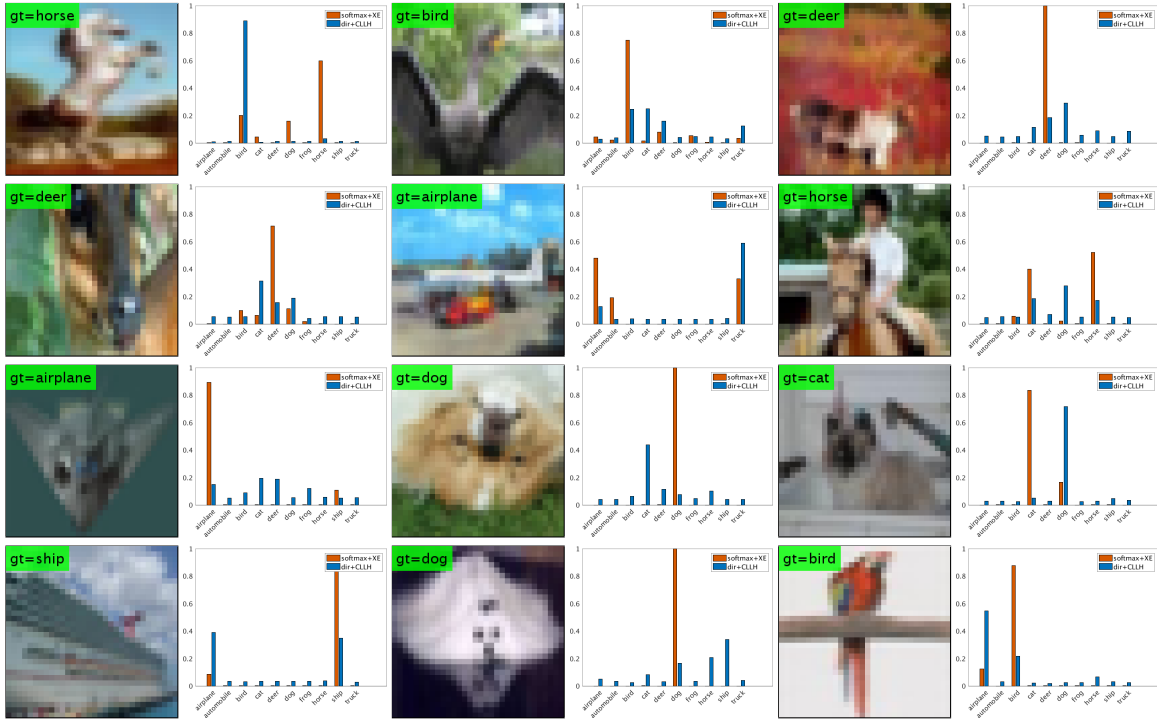

Figure 9. **Failure cases for Dirichlet layer**. Color coding as before. Here, the softmax predicts the correct classes, while the Dirichlet layer fails. However, in many cases the Dirichlet yields highly uncertain, *i.e.* high entropy, predictions.

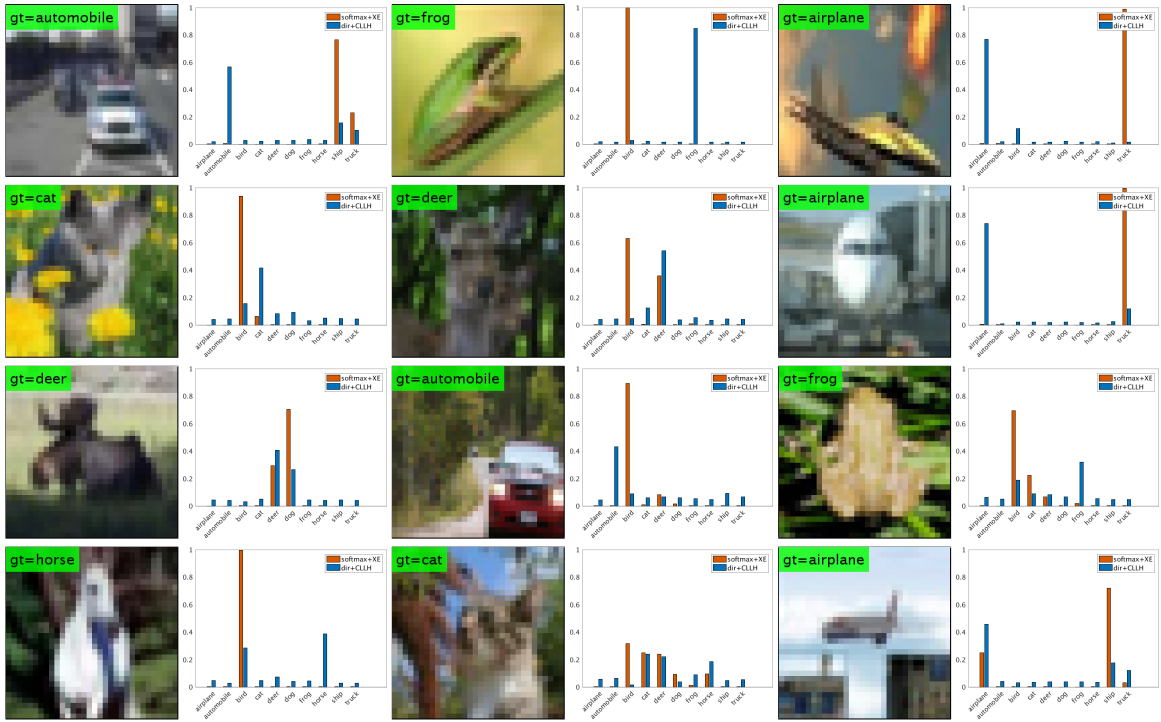

Figure 10. **Failure cases for softmax**. Color coding as before. Here, the softmax fails, while the Dirichlet layer succeeds. Note how confident softmax predictions tend to be, despite predicting the wrong class (*c.f.* airplanes top right).

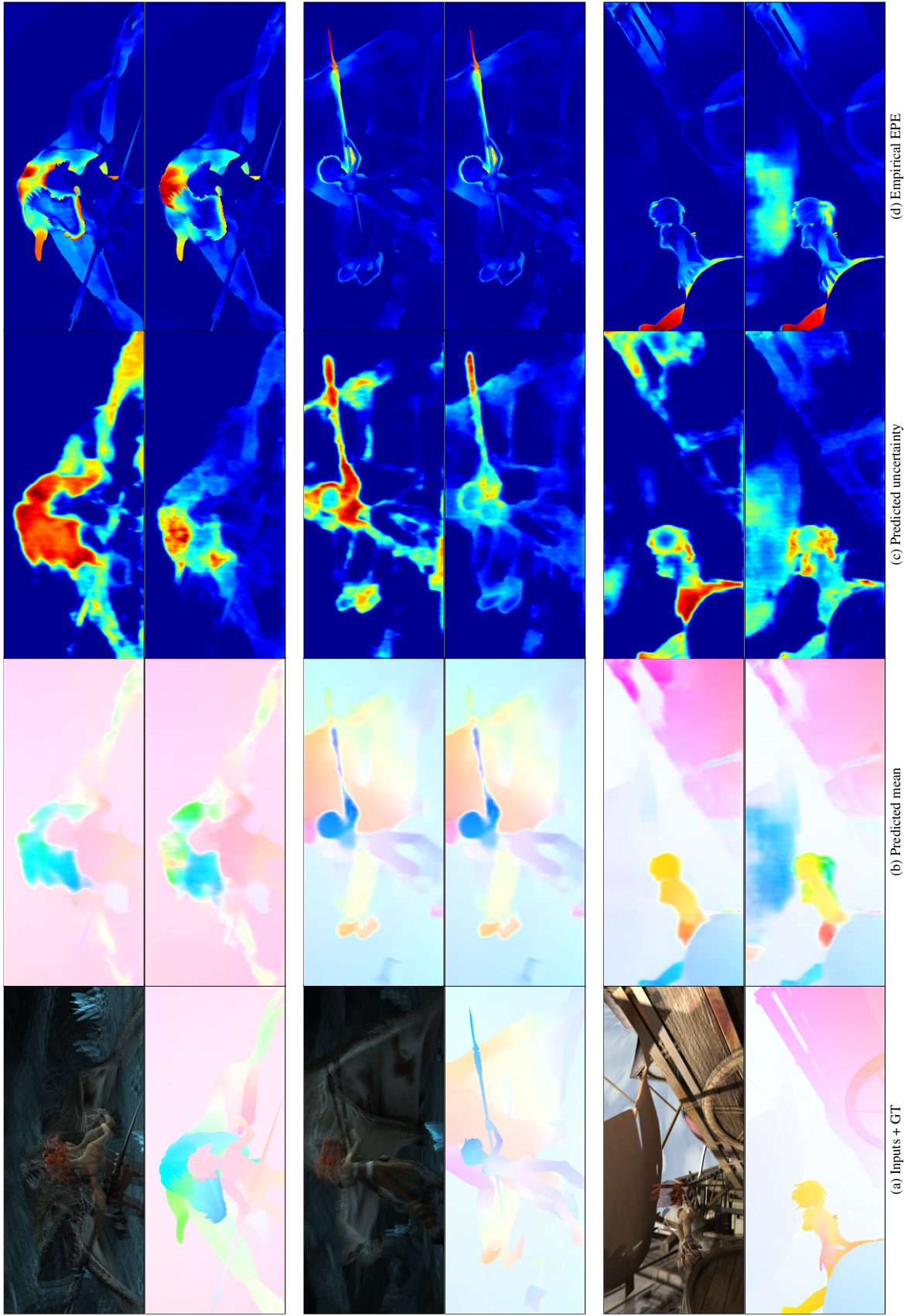

Figure 11. **Probabilistic regression of optical flow on Sintel images.** Our lightweight probabilistic CNNs, FlowNetADF (top of each block) and FlowNetProbOut (bottom of each block), yield uncertainties for predictions while staying competitive w.r.t. the endpoint error (EPE). The uncertainties are visibly correlated with the EPE.

(a) Inputs + GT  (b) Predicted mean  (c) Predicted uncertainty  (d) Empirical EPE

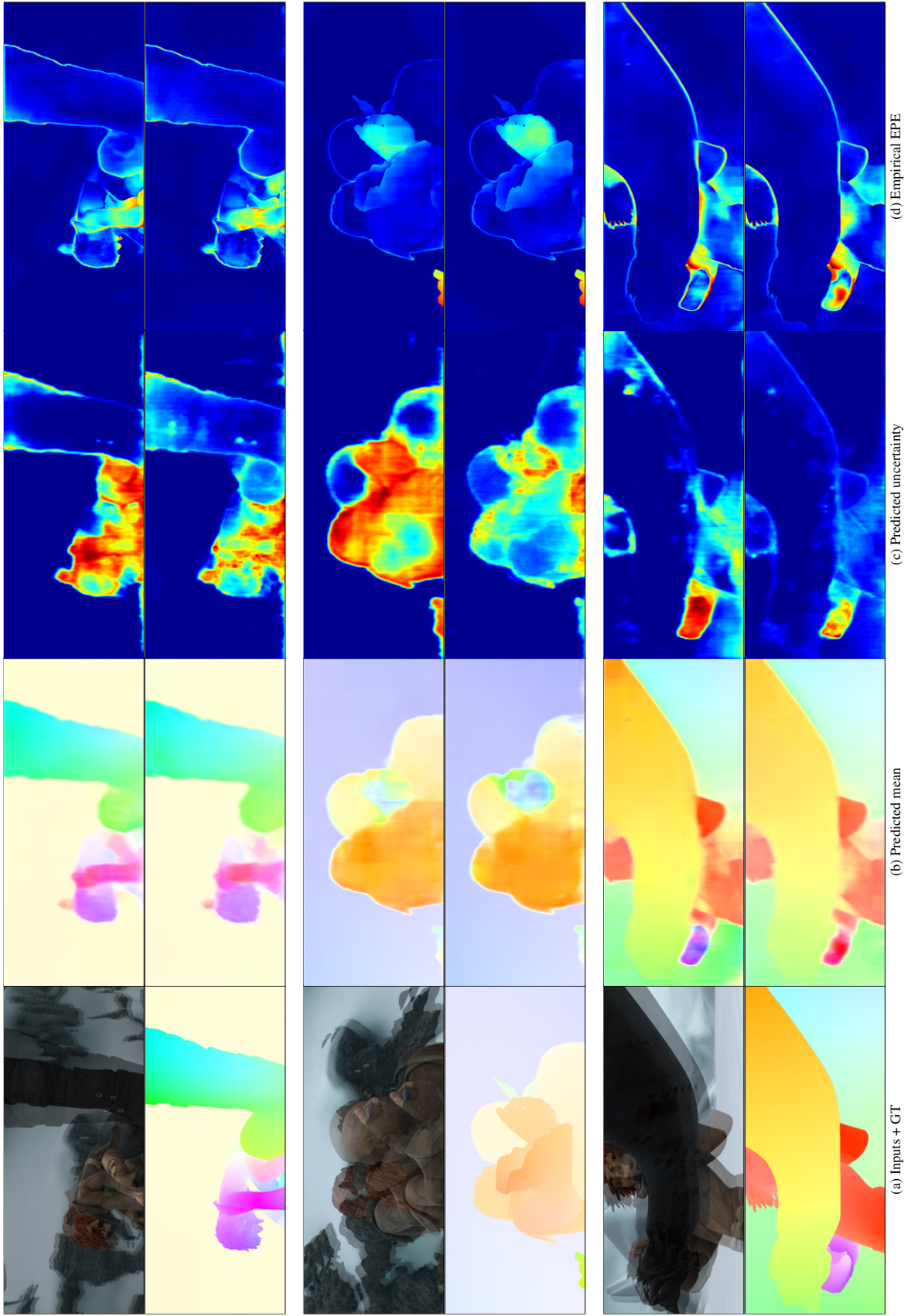

Figure 12. **Probabilistic regression of optical flow on Sintel images.** Our lightweight probabilistic CNNs, FlowNetADF (top of each block) and FlowNetProbOut (bottom of each block), yield uncertainties for predictions while staying competitive w.r.t. the endpoint error (EPE). The uncertainties are visibly correlated with the EPE.